\documentclass[11pt]{article}
\pdfoutput=1

\usepackage{times}
\usepackage{latexsym}
\usepackage{amsmath}
\usepackage{multirow}
\usepackage{url}
\usepackage{graphicx}
\usepackage[linesnumbered,ruled,vlined,commentsnumbered]{algorithm2e}
\SetAlFnt{\small}
\usepackage{amssymb}
\usepackage{amsfonts}
\usepackage{todonotes}
\usepackage[hidelinks]{hyperref}
\usepackage{fancyvrb}

\usepackage[T1,T5]{fontenc}

\usepackage {xcolor}
\usetikzlibrary {positioning}
\usepackage[inline]{enumitem}

\usepackage{acl2015}

\newcommand{\citep}[1]{\cite{#1}}
\newcommand{\citet}[1]{\newcite{#1}}
\newcommand{\viet}[1]{\textit{\fontencoding{T5}\selectfont #1}}

\title{An empirical study for Vietnamese dependency parsing\thanks{\ \ To appear in Proceedings of the Australasian Language Technology Association Workshop 2016.}}

\author{Dat Quoc Nguyen, Mark Dras \and Mark Johnson \\
Department of Computing \\ 
Macquarie University, Australia \\
{\tt {\footnotesize dat.nguyen@students.mq.edu.au, \{mark.dras, mark.johnson\}@mq.edu.au}}
}

\begin{document}
\maketitle

\begin{abstract}
This paper presents an empirical comparison of different dependency parsers for Vietnamese, which has some unusual characteristics such as copula drop and verb serialization. Experimental results show that the neural network-based parsers perform significantly better than the traditional parsers. We report the highest parsing scores published to date for Vietnamese with the labeled attachment score (LAS) at 73.53\% and the unlabeled attachment score (UAS) at 80.66\%.
\end{abstract}

\section{Introduction}

Dependency parsing has become a key research topic in natural language
processing in the last decade, boosted by the success of the CoNLL
2006 and 2007 shared tasks on multilingual dependency parsing
\cite{Buchholz2006,Nivre07}. \citet{McDonald2011Nivre} identify two
types of approaches for dependency parsing: graph-based approaches
\cite{McDonald2005OLT} and transition-based approaches
\cite{Nivre2007}.  Most traditional graph- or transition-based
dependency parsers
\cite{McDonald2005OLT,Nivre2007,Bohnet2010,zhang-nivre:2011:ACL-HLT2011,martinsalmeidasmith2013,choi-mccallum:2013:ACL2013}
manually define a set of core and combined features associated with
one-hot representations.

Recent work shows that neural network-based parsers obtain the
state-of-the-art parsing results across many
languages. \citet{chen-manning:2014:EMNLP2014},
\citet{weiss-EtAl:2015:ACL-IJCNLP},
\citet{pei-ge-chang:2015:ACL-IJCNLP}, and \citet{andor-EtAl:2016:P16-1}
represent the core features with dense vector embeddings and then feed
them as inputs to neural network-based classifiers, while
\citet{dyer-EtAl:2015:ACL-IJCNLP}, \citet{TACL798}, and \citet{TACL885}
propose novel neural network architectures to solve the
feature-engineering problem.

Dependency parsing for Vietnamese has not been actively explored. One
main reason is because there is no manually labeled dependency
treebank available. \citet{Thi2013} and \citet{Nguyen2014NLDB} propose
constituent-to-dependency conversion approaches to automatically
translate the manually built constituent treebank for Vietnamese
\citep{nguyen-EtAl:2009:LAW-III} to dependency treebanks. The
converted dependency treebanks are then used in later works on
Vietnamese dependency parsing, including \citet{Vu-Manh2015},
\citet{Le-Hong2015} and \citet{7371762}. All of the previous research
works use either the MSTparser \cite{McDonald2005OLT} or the
Maltparser \cite{Nivre2007} for their parsing experiments. Among them,
\citet{Nguyen2014NLDB} report the highest results with LAS at 71.66\%
and UAS at 79.08\% obtained by MSTparser.  However, MSTparser and
Maltparser are no longer considered state-of-the-art parsers.

In this paper, we present an empirical study of Vietnamese dependency
parsing.  We make comparisons between neural network-based parsers and
traditional parsers, and also between graph-based parsers and
transition-based parsers. We show that the neural network-based
parsers obtain significantly higher scores than the traditional
parsers. Specifically, we report the highest up-to-date scores for
Vietnamese with LAS at 73.53\% and UAS at 80.66\%. We also examine
potential problems specific to parsing Vietnamese, and point out
potential solutions for improving the parsing performance.

\setlength{\abovecaptionskip}{3pt plus 1pt minus 1pt} 
\begin{table}[t]
\centering
\begin{tabular}{l|l|l|l|l|l}  
\hline
\multicolumn{2}{c}{\bf Dep. labels} & \multicolumn{2}{|c}{\bf POS tags} & \multicolumn{2}{|c}{\bf Sent. length}\\
\hline
Type & Rate & Type & Rate & Length & Rate \\
\hline
adv	& 5.9 & A & 6.0 & $1 - 10$ & 19.0\\
amod	 & 2.4  & C & 3.7 & $11 - 20$ & 35.4 \\
conj	 & 1.9  & E & 6.5& $21 - 30$ & 25.6 \\
coord	& 1.9  & M & 3.6 & $31 - 40$ & 12.2 \\
dep	&  3.1  & N & 24.6 & $41 - 50$ & 4.9\\
det	& 6.2  & Nc & 2.4 & $> 50$ & 2.9 \\
dob	& 6.0  & Np & 4.2   & \_ & \_\\
loc	& 2.3  & P & 4.0  & \_ & \_\\
nmod	 & 19.0 & R & 7.4   & \_ & \_\\
pob	& 5.6  & V & 19.4   & \_ & \_\\
punct & 13.9  & \_ & \_  & \_ & \_\\
root	 & 4.7  &  \_ & \_  & \_ & \_\\
sub	& 6.8  &  \_ & \_  & \_ & \_\\
tmp	& 2.2 &  \_ & \_  & \_ & \_\\
vmod	 & 14.8 &  \_ & \_  & \_ & \_\\
\hline
\end{tabular}
\caption{VnDT statistics by  most frequent dependency  and part-of-speech (POS) labels,  and sentence length (i.e. number of words). ``Rate'' denotes the percentage occurrence in VnDT.  Dependency labels: \textit{adv} (adverbial), \textit{amod} (adjectival modifier), \textit{conj} (conjunct), \textit{coord} (coordinating conjunction), \textit{dep} (unspecified dependency), \textit{det} (determiner), \textit{dob} (direct object), \textit{loc} (location), \textit{nmod} (noun modifier), \textit{pob} (object of a preposition), \textit{punct} (punctuation), \textit{sub} (subject), \textit{tmp}  (temporal), \textit{vmod} (verb modifier). POS tags: A (Adjective), C (Conjunction), E (Preposition), M (Quantity), N (Noun), Nc (Classifier noun), Np (Proper noun), P (Pronoun), R (Adjunct), V (Verb).}
\label{tab:vndt}
\end{table}

\section{Experimental setup}

\paragraph{Dataset:}
There are two Vietnamese dependency treebanks which are automatically
converted from the manually-annotated Vietnamese constituent treebank
\citep{nguyen-EtAl:2009:LAW-III}, using conversion approaches proposed
by \citet{Thi2013} and \citet{Nguyen2014NLDB}. In \citet{Thi2013}'s
conversion approach, it is not clear how the dependency labels are
inferred; also, it ignores grammatical information encoded in
grammatical function tags. In addition, \citet{Thi2013}'s approach is
unable to handle cases of coordination and empty category mappings,
which frequently appear in the Vietnamese constituent treebank.
\citet{Nguyen2014NLDB} later proposed a new conversion approach to handle
those cases, with a better use of existing information in the
Vietnamese constituent treebank. So we conduct experiments using VnDT,
the high quality Vietnamese dependency treebank produced by
\citet{Nguyen2014NLDB}. The VnDT treebank consists of 10,200 sentences
(about 219K words). Table \ref{tab:vndt} gives some basic statistics
of VnDT. We use the last 1020 sentences of VnDT for testing while the
remaining sentences are used for training, resulting in an
out-of-vocabulary rate of 3.3\%.

\paragraph{Dependency parsers:} We experiment with four  parsers:   the graph-based parsers BIST-bmstparser\footnote{{\scriptsize\url{https://github.com/elikip/bist-parser/tree/master/bmstparser}}} (\textbf{BistG}) and MSTparser\footnote{{\scriptsize\url{http://www.seas.upenn.edu/~strctlrn/MSTParser/MSTParser.html}}} (\textbf{MST}), and the transition-based parsers BIST-barchybrid\footnote{{\scriptsize\url{https://github.com/elikip/bist-parser/tree/master/barchybrid}}}  (\textbf{BistT}) and Maltparser\footnote{{\scriptsize\url{http://www.maltparser.org}}} (\textbf{Malt}). The state-of-the-art BistG and BistT parsers \cite{TACL885} employ a bidirectional LSTM RNN architecture \cite{Schuster1997BRN,HochreiterSchmidhuber1997b} to automatically learn the feature representation.  In contrast, the traditional parsers MST \cite{McDonald2005OLT} and Malt \cite{Nivre2007}  use a  set of predefined features.  For training these parsers, we used the default settings.

\paragraph{Evaluation metrics:} The metrics are the labeled attachment score (LAS), unlabeled attachment score (UAS) and label accuracy score (LS). LAS is the percentage of words which are correctly assigned both dependency arc and label while UAS is the percentage of words for which the dependency arc is assigned correctly, and LS is the percentage of words for which the dependency label is assigned correctly.

\begin{table*}[!t]
\centering
\resizebox{16cm}{!}{
\begin{tabular}{ll|lll|lll|lll|lll}
\hline
\multicolumn{2}{c}{\multirow{3}{*}{\bf System}} & \multicolumn{6}{|c}{\bf With punctuation} & \multicolumn{6}{|c}{\bf Without punctuation}\\
\cline{3-14}
& & \multicolumn{3}{|c}{\bf Overall} & \multicolumn{3}{|c}{\bf Exact match} & \multicolumn{3}{|c}{\bf Overall} & \multicolumn{3}{|c}{\bf Exact match}\\
\cline{3-14}
\cline{3-14}
& &   LAS & UAS & LS &   LAS & UAS & LS &   LAS & UAS & LS &   LAS & UAS & LS \\
\hline

\multirow{4}{*}{\rotatebox[origin=c]{90}{Gold POS}} & BistG  & \textbf{73.17} & \textbf{79.39} & \textbf{84.22} & \textbf{11.27} & \textbf{19.71} & 15.20 & \textbf{73.53} & 80.66 & \textbf{81.86} & \textbf{11.96}	& \textbf{20.88}	& 15.20\\
& BistT  & 72.53 & 79.33 & 83.71 & \textbf{11.27} & 19.41 & \textbf{16.18} & 72.91 & \textbf{80.73} & 81.29 & 11.67	& 20.29	& \textbf{16.18}\\
& MST 	 & 70.29	 & 76.47 & 83.23 & 8.43	& 12.94	& 14.02 & 71.61 & 78.71 & 80.72 & 9.80	 & 16.37 & 	14.02\\
& Malt  & 69.10 & 74.91 & 81.72 & 9.22	& 14.80	& 13.92 & 70.39 & 77.08 & 79.33 & 9.71	 & 17.16 & 	13.92\\
\hline
\hline
\multirow{4}{*}{\rotatebox[origin=c]{90}{Auto POS}} & BistG  & \textbf{68.40} & 76.28 & \textbf{80.56} & 9.12	 & 16.18 & 11.76 & \textbf{68.50}	& 77.55	& \textbf{77.65} & 9.71 & 	\textbf{17.25} & 11.76\\
& BistT  & 68.22 & \textbf{76.56} &	80.22 & \textbf{9.80} & \textbf{16.27} & \textbf{13.24} & 68.31	& \textbf{77.91}	& 77.27 & 	\textbf{10.00}	& 17.06	& \textbf{13.24}\\
& MST 	 & 65.99	 & 73.94 & 79.78 & 6.86	& 10.78	& 10.88 & 66.99 & 76.12 & 76.75 & 7.84 & 	13.33	& 10.88\\ 
& Malt  & 64.94 & 72.32 & 78.43 & 7.35	& 12.25	& 10.20 & 65.88 & 74.36 & 75.56 & 7.55	 & 14.02 &	10.20\\
\hline
\end{tabular}
}
\caption{Parsing results. ``Without punctuation'' denotes parsing results where the punctuation and other symbols are excluded from evaluation. ``Exact match'' denotes the proportion of sentences whose predicted dependency trees are entirely correct.}
\label{tab:mainresults}
\label{tab:goldPOS}
\end{table*}

\section{Main results}

\subsection{Overall accuracy}

Table \ref{tab:mainresults} compares the parsing results obtained by the four  parsers. The first four rows report the scores with gold part-of-speech (POS) tags while the last four rows present the scores with automatically predicted POS tags.\footnote{We adapted the RDRPOSTagger toolkit  \protect{\cite{NguyenEACL2014NPP,NguyenNPP_AICom2015}} to automatically assign POS tags to words in the test set with an accuracy of 94.58\%.} 

As expected the neural network-based parsers BistG and BistT perform
significantly better than the traditional parsers MST and
Malt.\footnote{Using McNemar's test, the differences are statistically
  significant at $p <$ 0.001.} Specifically, we find 2$^+$\% absolute
improvements in LAS and UAS scores in both graph- and transition-based
types. In most cases, there are no significant differences between the
LAS and UAS scores of BistG and BistT, except LAS scored on gold POS
tags (73.17\% against 72.53\%, and 73.53\% against
72.91\%).\footnote{The differences are statistically significant at $p
  <$ 0.02.} Compared to the previous highest results (LAS at 71.66\%
and UAS at 79.08\%) scored without punctuation on gold POS tags in
\citet{Nguyen2014NLDB}, we obtain better scores (LAS at 73.53\% and
UAS at 80.66\%) with BistG.

Next, Section \ref{sec:aa} gives a detailed accuracy analysis on gold POS tags \textbf{without} punctuation, and Section \ref{sec:ea} discusses the source of some errors and possible improvements.

\subsection{Accuracy analysis}
\label{sec:aa}

\begin{figure}[t]
\includegraphics[width=7cm]{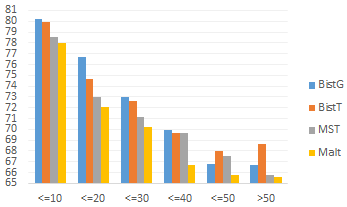}
\caption{LAS  by sentence length.}
\label{fig:lasSL}
\end{figure}

\begin{figure}[t]
\includegraphics[width=7cm]{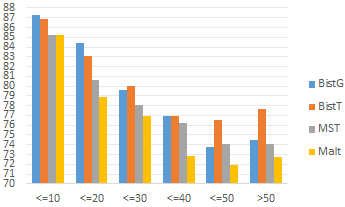}
\caption{UAS  by sentence length}
\label{fig:uasSL}
\end{figure}

\paragraph{Sentence length:} Figures \ref{fig:lasSL} and \ref{fig:uasSL} detail LAS and UAS scores by sentence length in bins of length 10. It is not surprising that all parsers produce better results for shorter sentences. For sentences shorter than 10 words, all LAS and UAS scores are around 80\% and 85\%, respectively. However, the scores drop by 10$^+$\% for sentences longer than 50 words. The Malt parser obtains the lowest LAS and UAS scores across all sentence bins. BistG obtains the highest scores for sentences shorter than 20 words while BistT obtains highest scores for sentences longer than 40 words. BistG, BistT and MST perform similarly on 30-to-40-word sentences. For shorter sentences from 20 to 30 words, BistG and BistT produce similar results but higher than obtained by MST.

\begin{figure}[t]
\includegraphics[width=7cm]{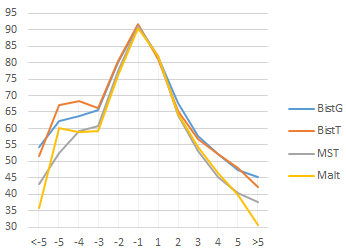}
\caption{F$_1$ scores by dependency distance for labeled attachment}
\label{fig:lasdd}
\end{figure}

\begin{figure}[t]
\includegraphics[width=7cm]{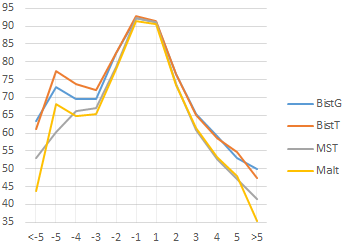}
\caption{F$_1$ scores  by dependency distance for unlabeled attachment}
\label{fig:uasdd}
\end{figure}

\paragraph{Dependency distance:} Figures \ref{fig:lasdd} and \ref{fig:uasdd} show F$_1$ scores in terms of the distance from each dependent word to its head. Similar to English \cite{choitetreaultstent2015}, we find better predictions for the left dependencies than for the right dependencies. Unlike in English where the lower scores are associated with longer distances, we find a different pattern when predicting the left dependencies in Vietnamese. 
In a distance bin of 3, 4 and 5 words with respect to the left dependencies, three over four parsers including BistG, BistT and Malt generally obtain better predictions for longer distances. Compared to English, Vietnamese is head-initial, so finding a difference with respect to left dependencies is not completely unexpected. In addition, for this distance bin, the transition-based parser does better than the graph-based parser in both neural net-based and traditional categories (i.e. BistT $>$ BistG and Malt $>$ MST).  In both those categories, however, the graph-based parser does better than the transition-based parser for 5-word-longer distances (i.e. BistG $>$ BistT and MST $>$ Malt), while they produce similar results on dependency distances of 1 or 2 words. 

Because the dependency distance of 3, 4 or 5 occurs quite frequently in long sentences, so the results here are consistent with the results shown in Figures \ref{fig:lasSL} and \ref{fig:uasSL} where BistT obtains the highest scores for long sentences.

\paragraph{Dependency labels:} Table \ref{tab:dl} presents LAS scores for the most frequent dependency labels. The labels with higher than 90\% accuracy are \emph{adv}, \emph{det} and \emph{pob}  in which surprisingly MST obtains the best results on all these labels and even on both \emph{conj} and \emph{dob} labels. BistT obtains the best scores on the two most frequent labels \emph{nmod} and \emph{vmod}, and also on the \emph{loc} label. BistG performs best on the remaining labels. Biggest ranges ($> 10\%$) of obtained scores across parsers associate to  labels \emph{coord}, \emph{dep}, \emph{sub} and \emph{tmp}.

\begin{table}[!ht]
\centering
\resizebox{7.5cm}{!}{
\begin{tabular}{l||l|l|l|l||l}  
\hline
Type & BistG & 	BistT & MST & Malt & Avg. \\ 
\hline
adv	 & 92.09	& \textbf{92.40}	& \textbf{92.40}	& 92.33	& 92.31 \\ 
amod	  & \textbf{77.30}	& 73.89	& 76.11	& 73.21	& 75.13 \\ 
conj	  & 74.82	& 73.11	& \textbf{78	.00}& 71.64	& 74.39 \\ 
coord  & \textbf{57.49}	& 49.52	& 46.14	& 52.66	& 51.45 \\ 
dep	 & \textbf{47.83}	& 46	.00& 32.54	& 42.08	& 42.11 \\ 
det	 & 94.15	& 94.30	& \textbf{95.27}	& 94.52 & 94.56\\ 
dob	 & 73.01	& 70.81	& \textbf{78.62}	& 76.35	& 74.70 \\ 
loc	 & 52.54	& \textbf{53.86}	& 51.43	& 50.77	 & 52.15 \\ 
nmod	  & 79.34	& \textbf{79.51}	& 78.10	& 76.67	& 78.41 \\ 
pob	 & 94.35	& 95.27	& \textbf{96.18}	& 95.85	& 95.41 \\ 
root	  & \textbf{85.69}	& 82.55	& 82.06	& 74.41	& 81.18 \\ 
sub	 & \textbf{73.34}	& 72.61	& 66.49	& 62.67	& 68.78 \\ 
tmp	  & \textbf{60.68}	& 57.05	& 44.66	& 41.45 & 50.96 \\ 
vmod	  & 61.51	& \textbf{62.02}	& 60.79	& 60.23	& 61.14 \\ 
\hline
\end{tabular}
}
\caption{LAS  by  most frequent dependency labels. ``Avg.'' denotes the averaged score of four parsers.}
\label{tab:dl}
\end{table}

\begin{table}[!ht]
\resizebox{7.75cm}{!}{
\setlength{\tabcolsep}{0.25em}
\begin{tabular}{l||l|l|l|l|l|l|l|l}
\hline
\multirow{2}{*}{\bf POS} & \multicolumn{4}{|c}{\bf LAS}  & \multicolumn{4}{|c}{\bf UAS} \\
\cline{2-9}
& BistG & 	BistT & MST & Malt & BistG & 	BistT & MST & Malt\\
\hline
A  & 68.32		& \textbf{70.31}		& 69.89		& 66.83  & 73.01	& \textbf{75.50}	& 74.86	& 70.88 \\
C 	& \textbf{55.90}	 	& 50.00		& 44.94		& 50.00  & \textbf{61.33}	& 56.87	& 50.60	& 54.94 \\
E	 & \textbf{55.47	}	& 53.87		& 50.91		& 49.96 & \textbf{72.27}	& 71.54	& 68.86	& 64.45 \\
M	 & 92.11		& 91.05		& \textbf{93.03}		& 91.18 & 93.42	& 93.16	& \textbf{94.21}	& 91.71 \\
N	 & \textbf{74.37}		& 73.58		& 73.77		& 71.30 & \textbf{83.95}	& 83.86	& 82.58	& 80.48 \\
Nc	 & 69.86		& \textbf{72.02}		& 68.49		& 67.12 & 78.47	& \textbf{79.26}	& 76.13	& 74.17 \\
Np	 & 84.69		& 84.47		& \textbf{84.80}		& 82.84 & \textbf{88.49}	& \textbf{88.49}	& 88.06	& 86.43 \\
P	 & 79.34		& \textbf{80.16}		& 79.69		& 77.23 & 85.45	& \textbf{85.92}	& 84.62	& 82.39 \\
R	 & 91.94		& \textbf{93.08}		& 92.42		& 92.60 & 92.90	& \textbf{93.87}	& 92.96	& 93.27 \\
V	 & \textbf{68.01}		& 66.49		& 63.78		& 63.13 & \textbf{75.05}	& 74.95	& 71.83	& 70.49 \\
\hline
\end{tabular}
}
\caption{Results by most frequent POS tags.}
\label{tab:laspos}
\end{table}

Table \ref{tab:dl} also shows that the label with the lowest LAS scores ($<50\%$) across all parsers is \emph{dep} which is a very general label. Those with LAS scores ranging from  $50\%$ to about $60\%$ are \emph{coord}, \emph{loc}, \emph{tmp} and \emph{vmod} in which \emph{coord}, \emph{loc} and \emph{tmp} are among the least frequent labels, while \emph{vmod} is the second most frequent label. 

\paragraph{POS tags:} In Table \ref{tab:laspos} we  analyze the results by the POS tag of the dependent. BistG achieves the highest results on the two most frequent POS tags \emph{N}   and \emph{V}   and also on \emph{C}   and \emph{E}. BistT achieves the highest scores on the remaining POS tags except \emph{M}   for which MST produces the highest score.

\subsection{Discussions}
\label{sec:ea}

\paragraph{Linguistic aspects:}
One surprising characteristic of the results is the poor performance
of verb-related dependencies: \emph{vmod} accuracy is low, as are
scores associated with the second most frequent POS tag \emph{V}
(Verb). For the latter, we find significantly lower scores for verbs
in Vietnamese (around 65\% as shown in Table \ref{tab:laspos}) against
scores for verbs (about 80$^+$\%) obtained by MST and Malt parsers on
13 other languages reported in \citet{McDonald2011Nivre}, and also
much worse performance in terms of rank relative to other POS.

\begin{figure}[t]
 \centering
\includegraphics[width=6cm]{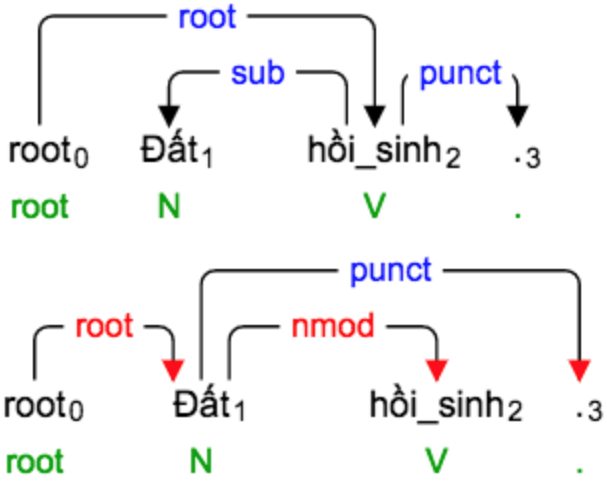}
\caption{An example  of a Vietnamese sentence with copula-drop. The first parsed tree is the gold one  while the second parsed tree is the same output produced by all parsers. This sentence is translated to English as ``{the land$_{\#1}$ is revived$_{\#2}$ .}'', in which the copula ``is'' is dropped in  Vietnamese. The subscripts in the English-translated sentence refer to  alignments with the word indexes in the Vietnamese sentence.}
\label{fig:error1}
\end{figure}

This may be related to syntactic characteristics of Vietnamese
\citep{Thompson1987}.  First, Vietnamese is described as a copula-drop
language.  Consider \viet{C\^o H\`a c\'o nh\`a \dj\d ep} ``Miss H\`a
has a beautiful house'', where the attributive adjective \viet{\dj\d
  ep} ``beautiful'' postmodifies the noun \viet{nh\`a} ``house''.
Adjectives can also be predicative, where they are conventionally
labelled \emph{V} (Verb), and a copula is absent: with Vietnamese's
SVO word order, this is also \viet{nh\`a \dj\d ep} ``the house is
beautiful.''  Figure~\ref{fig:error1} presents an example from the treebank:
all four parsers produce the incorrect structure, which is what would
be expected for the attributive adjectival use in an NP.  This
construction is quite common in Vietnamese.

\begin{figure}[t]
 \centering
\includegraphics[width=7.5cm]{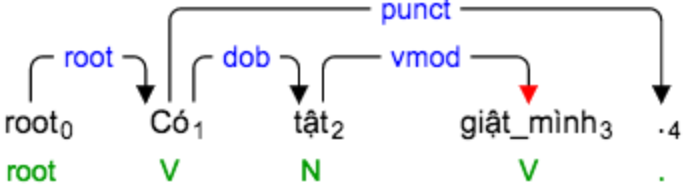}
\caption{An example  of a Vietnamese sentence with verb serialization (and pronoun-dropping), parsed by BistT. The gold parsed tree is when the indexed-3 word is attached to the indexed-1 word by \textit{vmod}, instead of the indexed-2 word. For this sentence, BistG, MST and Malt attach  the indexed-3 word to be dependent to the indexed-2 word by the label \textit{nmod}.  This  sentence is translated to English as ``{He who excuses$_{\#1}$ himself$_{\#2}$, accuses$_{\#3}$ himself .}'' }
\label{fig:error}
\end{figure}

Second, Vietnamese permits verb serialization, as in Figure
\ref{fig:error}: \viet{gi\d \acircumflex{}t\_m\`inh} ``accuses'' should be
a \emph{vmod} dependent of \viet{c\'o} ``excuses''; such a construction
is analogous to the more familiar \emph{nmod} in other languages.
Verb dependencies in Vietnamese might thus be less predictable than in
other languages, with a more varied distribution of dependents.

\paragraph{Other aspects:}
Generally, one reason for low overall scores on Vietnamese dependency
parsing when compared to the scores obtained on the other languages
\citep{McDonald2011Nivre} is probably because of the complex
structures of many long sentences in the VnDT treebank (e.g. 45\% of
the sentences in VnDT consist of more than 20 words).  So we can only
obtain 60\% and 50\% for left and right dependency distances larger
than 5 as shown in Figure \ref{fig:uasdd}, respectively, while for
English both left and right dependencies with distances larger than 5
have greater than 70\% accuracy \cite{choitetreaultstent2015}.

\begin{table}[t]
\resizebox{7.5cm}{!}{
\begin{tabular}{l||l|l|l|l|l|l}
\hline
\multirow{2}{*}{\bf Oracle} & \multicolumn{3}{|c}{\bf With punct.}  & \multicolumn{3}{|c}{\bf Without punct.} \\
\cline{2-7}
& LAS & UAS & LS & LAS & UAS & LS\\
\hline
Tree	& 79.20	& 85.22	& 88.38 & 79.33	& 86.24	& 86.66\\
Arc	& 85.98	& 90.50	& 92.67 & 85.96	& 91.14	& 91.57\\
\hline
\end{tabular}
}
\caption{Upper bound of  ensemble performance.}
\label{tab:ensembles}
\end{table}

One simple approach to improve parsing performance for Vietnamese is
to separately use the graph-based parser BistG for short sentences and
the transition-based parser BistT for longer sentences. Another
approach is to use system combination
\cite{nivre-mcdonald:2008:ACLMain,zhang-clark:2008:EMNLP},
e.g. building ensemble systems
\citep{sagaetsujii2007,Surdeanu2010EMD,haffarirazavisarkar2011}. Table
\ref{tab:ensembles} presents an \underline{upper bound} of oracle
ensemble performance, using the \textsc{dependable} toolkit
\citep{choitetreaultstent2015}. \textsc{dependable} assumes that
either the best tree or the best arc can be determined by an oracle.

\section{Conclusions}

We have presented  an empirical comparison for Vietnamese dependency parsing. Experimental results on the Vietnamese dependency treebank VnDT \cite{Nguyen2014NLDB} show that the neural network-based parsers \cite{TACL885} obtain significantly higher scores than the traditional parsers \cite{McDonald2005OLT,Nivre2007}. More specifically, in each graph- or transition-based type, we find a 2\% absolute improvement of the neural network-based parser over the traditional one.  

 We report the highest  performance up to date for Vietnamese dependency parsing with LAS at 73.53\% and UAS at 80.66\%.

\section*{Acknowledgments}
The first author is supported by an International Postgraduate Research Scholarship and a NICTA NRPA Top-Up Scholarship. 

{
\hyphenpenalty=10000
\bibliographystyle{acl}
\bibliography{References}

\begin{thebibliography}{}

\bibitem[\protect\citename{Andor \bgroup et al.\egroup
  }2016]{andor-EtAl:2016:P16-1}
Daniel Andor, Chris Alberti, David Weiss, Aliaksei Severyn, Alessandro Presta,
  Kuzman Ganchev, Slav Petrov, and Michael Collins.
\newblock 2016.
\newblock {Globally Normalized Transition-Based Neural Networks}.
\newblock In {\em Proceedings of the 54th Annual Meeting of the Association for
  Computational Linguistics}, pages 2442--2452.

\bibitem[\protect\citename{Bohnet}2010]{Bohnet2010}
Bernd Bohnet.
\newblock 2010.
\newblock Very high accuracy and fast dependency parsing is not a
  contradiction.
\newblock In {\em Proceedings of the 23rd International Conference on
  Computational Linguistics}, pages 89--97.

\bibitem[\protect\citename{Buchholz and Marsi}2006]{Buchholz2006}
Sabine Buchholz and Erwin Marsi.
\newblock 2006.
\newblock {CoNLL-X shared task on multilingual dependency parsing}.
\newblock In {\em Proceedings of the Tenth Conference on Computational Natural
  Language Learning, CoNLL-X}, pages 149--164.

\bibitem[\protect\citename{Chen and Manning}2014]{chen-manning:2014:EMNLP2014}
Danqi Chen and Christopher Manning.
\newblock 2014.
\newblock A fast and accurate dependency parser using neural networks.
\newblock In {\em Proceedings of the 2014 Conference on Empirical Methods in
  Natural Language Processing}, pages 740--750.

\bibitem[\protect\citename{Choi and McCallum}2013]{choi-mccallum:2013:ACL2013}
Jinho~D. Choi and Andrew McCallum.
\newblock 2013.
\newblock Transition-based dependency parsing with selectional branching.
\newblock In {\em Proceedings of the 51st Annual Meeting of the Association for
  Computational Linguistics}, pages 1052--1062.

\bibitem[\protect\citename{Choi \bgroup et al.\egroup
  }2015]{choitetreaultstent2015}
Jinho~D. Choi, Joel Tetreault, and Amanda Stent.
\newblock 2015.
\newblock {It Depends: Dependency Parser Comparison Using A Web-based
  Evaluation Tool}.
\newblock In {\em Proceedings of the 53rd Annual Meeting of the Association for
  Computational Linguistics and the 7th International Joint Conference on
  Natural Language Processing}, pages 387--396.

\bibitem[\protect\citename{Dyer \bgroup et al.\egroup
  }2015]{dyer-EtAl:2015:ACL-IJCNLP}
Chris Dyer, Miguel Ballesteros, Wang Ling, Austin Matthews, and Noah~A. Smith.
\newblock 2015.
\newblock {Transition-Based Dependency Parsing with Stack Long Short-Term
  Memory}.
\newblock In {\em Proceedings of the 53rd Annual Meeting of the Association for
  Computational Linguistics and the 7th International Joint Conference on
  Natural Language Processing}, pages 334--343.

\bibitem[\protect\citename{Haffari \bgroup et al.\egroup
  }2011]{haffarirazavisarkar2011}
Gholamreza Haffari, Marzieh Razavi, and Anoop Sarkar.
\newblock 2011.
\newblock An ensemble model that combines syntactic and semantic clustering for
  discriminative dependency parsing.
\newblock In {\em Proceedings of the 49th Annual Meeting of the Association for
  Computational Linguistics: Human Language Technologies}, pages 710--714.

\bibitem[\protect\citename{Hochreiter and
  Schmidhuber}1997]{HochreiterSchmidhuber1997b}
Sepp Hochreiter and J{\"{u}}rgen Schmidhuber.
\newblock 1997.
\newblock Long short-term memory.
\newblock {\em Neural Computation}, 9(8):1735--1780.

\bibitem[\protect\citename{Kiperwasser and Goldberg}2016a]{TACL798}
Eliyahu Kiperwasser and Yoav Goldberg.
\newblock 2016a.
\newblock Easy-first dependency parsing with hierarchical tree lstms.
\newblock {\em Transactions of the Association for Computational Linguistics},
  4:445--461.

\bibitem[\protect\citename{Kiperwasser and Goldberg}2016b]{TACL885}
Eliyahu Kiperwasser and Yoav Goldberg.
\newblock 2016b.
\newblock {Simple and Accurate Dependency Parsing Using Bidirectional LSTM
  Feature Representations}.
\newblock {\em Transactions of the Association for Computational Linguistics},
  4:313--327.

\bibitem[\protect\citename{Le-Hong \bgroup et al.\egroup }2015]{Le-Hong2015}
Phuong Le-Hong, Thi-Minh-Huyen Nguyen, Thi-Luong Nguyen, and My-Linh Ha.
\newblock 2015.
\newblock Fast dependency parsing using distributed word representations.
\newblock In {\em Proceedings of the PAKDD 2015 Workshops}, pages 261--272.

\bibitem[\protect\citename{Martins \bgroup et al.\egroup
  }2013]{martinsalmeidasmith2013}
Andre Martins, Miguel Almeida, and Noah~A. Smith.
\newblock 2013.
\newblock {Turning on the Turbo: Fast Third-Order Non-Projective Turbo
  Parsers}.
\newblock In {\em Proceedings of the 51st Annual Meeting of the Association for
  Computational Linguistics (Volume 2: Short Papers)}, pages 617--622.

\bibitem[\protect\citename{McDonald and Nivre}2011]{McDonald2011Nivre}
Ryan McDonald and Joakim Nivre.
\newblock 2011.
\newblock Analyzing and integrating dependency parsers.
\newblock {\em Computational Linguistics}, 37(1):197--230.

\bibitem[\protect\citename{McDonald \bgroup et al.\egroup
  }2005]{McDonald2005OLT}
Ryan McDonald, Koby Crammer, and Fernando Pereira.
\newblock 2005.
\newblock Online large-margin training of dependency parsers.
\newblock In {\em Proceedings of the 43rd Annual Meeting on Association for
  Computational Linguistics}, pages 91--98.

\bibitem[\protect\citename{Nguyen and Nguyen}2015]{7371762}
Kiet~Van Nguyen and Ngan Luu-Thuy Nguyen.
\newblock 2015.
\newblock {Error Analysis for Vietnamese Dependency Parsing}.
\newblock In {\em Proceedings of the 2015 Seventh International Conference on
  Knowledge and Systems Engineering}, pages 79--84.

\bibitem[\protect\citename{Nguyen \bgroup et al.\egroup
  }2009]{nguyen-EtAl:2009:LAW-III}
Phuong~Thai Nguyen, Xuan~Luong Vu, Thi Minh~Huyen Nguyen, Van~Hiep Nguyen, and
  Hong~Phuong Le.
\newblock 2009.
\newblock {Building a Large Syntactically-Annotated Corpus of Vietnamese}.
\newblock In {\em Proceedings of the Third Linguistic Annotation Workshop},
  pages 182--185.

\bibitem[\protect\citename{Nguyen \bgroup et al.\egroup
  }2014a]{NguyenEACL2014NPP}
Dat~Quoc Nguyen, Dai~Quoc Nguyen, Dang~Duc Pham, and Son~Bao Pham.
\newblock 2014a.
\newblock {RDRPOSTagger: A Ripple Down Rules-based Part-Of-Speech Tagger}.
\newblock In {\em Proceedings of the Demonstrations at the 14th Conference of
  the European Chapter of the Association for Computational Linguistics}, pages
  17--20.

\bibitem[\protect\citename{Nguyen \bgroup et al.\egroup }2014b]{Nguyen2014NLDB}
Dat~Quoc Nguyen, Dai~Quoc Nguyen, Son~Bao Pham, Phuong-Thai Nguyen, and Minh~Le
  Nguyen.
\newblock 2014b.
\newblock {From Treebank Conversion to Automatic Dependency Parsing for
  Vietnamese}.
\newblock In {\em {Proceedings of 19th International Conference on Application
  of Natural Language to Information Systems}}, pages 196--207.

\bibitem[\protect\citename{Nguyen \bgroup et al.\egroup
  }2016]{NguyenNPP_AICom2015}
Dat~Quoc Nguyen, Dai~Quoc Nguyen, Dang~Duc Pham, and Son~Bao Pham.
\newblock 2016.
\newblock {A robust transformation-based learning approach using ripple down
  rules for part-of-speech tagging}.
\newblock {\em AI Communications}, 29(3):409--422.

\bibitem[\protect\citename{Nivre and
  McDonald}2008]{nivre-mcdonald:2008:ACLMain}
Joakim Nivre and Ryan McDonald.
\newblock 2008.
\newblock {Integrating Graph-Based and Transition-Based Dependency Parsers}.
\newblock In {\em Proceedings of ACL-08: HLT}, pages 950--958.

\bibitem[\protect\citename{Nivre \bgroup et al.\egroup }2007a]{Nivre07}
Joakim Nivre, Johan Hall, Sandra K\"{u}bler, Ryan McDonald, Jens Nilsson,
  Sebastian Riedel, and Deniz Yuret.
\newblock 2007a.
\newblock {The CoNLL 2007 Shared Task on Dependency Parsing}.
\newblock In {\em Proceedings of the CoNLL Shared Task Session of EMNLP-CoNLL
  2007}, pages 915--932.

\bibitem[\protect\citename{Nivre \bgroup et al.\egroup }2007b]{Nivre2007}
Joakim Nivre, Johan Hall, Jens Nilsson, Atanas Chanev, G\"{u}lsen Eryigit,
  Sandra K\"{u}bler, Svetoslav Marinov, and Erwin Marsi.
\newblock 2007b.
\newblock {MaltParser: A language-independent system for data-driven dependency
  parsing}.
\newblock {\em Natural Language Engineering}, 13(January):1.

\bibitem[\protect\citename{Pei \bgroup et al.\egroup
  }2015]{pei-ge-chang:2015:ACL-IJCNLP}
Wenzhe Pei, Tao Ge, and Baobao Chang.
\newblock 2015.
\newblock An effective neural network model for graph-based dependency parsing.
\newblock In {\em Proceedings of the 53rd Annual Meeting of the Association for
  Computational Linguistics and the 7th International Joint Conference on
  Natural Language Processing}, pages 313--322.

\bibitem[\protect\citename{Sagae and Tsujii}2007]{sagaetsujii2007}
Kenji Sagae and Jun'ichi Tsujii.
\newblock 2007.
\newblock Dependency parsing and domain adaptation with {LR} models and parser
  ensembles.
\newblock In {\em Proceedings of the CoNLL Shared Task Session of EMNLP-CoNLL
  2007}, pages 1044--1050.

\bibitem[\protect\citename{Schuster and Paliwal}1997]{Schuster1997BRN}
M.~Schuster and K.K. Paliwal.
\newblock 1997.
\newblock Bidirectional recurrent neural networks.
\newblock {\em IEEE Transactions on Signal Processing}, 45(11):2673--2681.

\bibitem[\protect\citename{Surdeanu and Manning}2010]{Surdeanu2010EMD}
Mihai Surdeanu and Christopher~D. Manning.
\newblock 2010.
\newblock Ensemble models for dependency parsing: Cheap and good?
\newblock In {\em Human Language Technologies: The 2010 Annual Conference of
  the North American Chapter of the Association for Computational Linguistics},
  pages 649--652.

\bibitem[\protect\citename{Thi \bgroup et al.\egroup }2013]{Thi2013}
Luong~Nguyen Thi, Linh~Ha My, Hung~Nguyen Viet, Huyen Nguyen~Thi Minh, and
  Phuong~Le Hong.
\newblock 2013.
\newblock {Building a treebank for Vietnamese dependency parsing}.
\newblock In {\em Proceedings of 2013 IEEE RIVF International Conference on
  Computing and Communication Technologies, Research, Innovation, and Vision
  for the Future}, pages 147--151.

\bibitem[\protect\citename{Thompson}1987]{Thompson1987}
Laurence~C. Thompson.
\newblock 1987.
\newblock {\em {A Vietnamese Reference Grammar}}.
\newblock University of Hawaii Press.

\bibitem[\protect\citename{Vu-Manh \bgroup et al.\egroup }2015]{Vu-Manh2015}
Cam Vu-Manh, Anh~Tuan Luong, and Phuong Le-Hong.
\newblock 2015.
\newblock Improving vietnamese dependency parsing using distributed word
  representations.
\newblock In {\em Proceedings of the Sixth International Symposium on
  Information and Communication Technology}, pages 54--60.

\bibitem[\protect\citename{Weiss \bgroup et al.\egroup
  }2015]{weiss-EtAl:2015:ACL-IJCNLP}
David Weiss, Chris Alberti, Michael Collins, and Slav Petrov.
\newblock 2015.
\newblock Structured training for neural network transition-based parsing.
\newblock In {\em Proceedings of the 53rd Annual Meeting of the Association for
  Computational Linguistics and the 7th International Joint Conference on
  Natural Language Processing}, pages 323--333.

\bibitem[\protect\citename{Zhang and Clark}2008]{zhang-clark:2008:EMNLP}
Yue Zhang and Stephen Clark.
\newblock 2008.
\newblock {A Tale of Two Parsers: Investigating and Combining Graph-based and
  Transition-based Dependency Parsing}.
\newblock In {\em Proceedings of the 2008 Conference on Empirical Methods in
  Natural Language Processing}, pages 562--571.

\bibitem[\protect\citename{Zhang and Nivre}2011]{zhang-nivre:2011:ACL-HLT2011}
Yue Zhang and Joakim Nivre.
\newblock 2011.
\newblock Transition-based dependency parsing with rich non-local features.
\newblock In {\em Proceedings of the 49th Annual Meeting of the Association for
  Computational Linguistics: Human Language Technologies}, pages 188--193.

\end{thebibliography}
}

\end{document}